\documentclass[runningheads]{llncs}
\usepackage{graphicx}
\usepackage{amsmath,amssymb} 
\usepackage{color}
\usepackage{times}
\usepackage{epsfig}
\usepackage{multirow}
\usepackage{makecell}
\begin{document}
\pagestyle{headings}
\mainmatter


\title{RAF-AU Database: In-the-Wild Facial Expressions with Subjective Emotion Judgement and Objective AU Annotations} 
\titlerunning{RAF-AU Database}
%
\author{Wen-Jing Yan\inst{1}\thanks{These authors contributed equally to this work.}
\and
Shan Li\inst{2 \star}
\and Chengtao Que\inst{1}\and Jiquan Pei\inst{1}\and Weihong Deng\inst{2}}
\authorrunning{W. J. Yan et al.}
%
\institute{JD Digits, Beijing, China\and
Beijing University of Posts and Telecommunications
\\
\email{eagan-ywj@foxmail.com,\{ls1995,whdeng\}@bupt.edu.cn,\\\{quechengtao,peijiquan\}@jd.com},}

\maketitle

\begin{abstract}
Much of the work on automatic facial expression recognition relies on databases containing a certain number of emotion classes and their exaggerated facial configurations (generally six prototypical facial expressions), based on Ekman's Basic Emotion Theory. However, recent studies have revealed that facial expressions in our human life can be blended with multiple basic emotions. And the emotion labels for these in-the-wild  facial expressions cannot easily be annotated solely on pre-defined AU patterns. How to analyze the action units for such complex expressions is still an open question. 
To address this issue, we develop a RAF-AU database that employs a sign-based (i.e., AUs) and judgement-based (i.e., perceived emotion) approach to annotating blended facial expressions in the wild. 
We first reviewed the annotation methods in existing databases and identified crowdsourcing as a promising strategy for labeling in-the-wild facial expressions. Then, RAF-AU was finely annotated by experienced coders, on which we also conducted a preliminary investigation of which key AUs contribute most to a perceived emotion, and the relationship between AUs and facial expressions.
Finally, we provided a baseline for AU recognition in RAF-AU using popular features and multi-label learning methods.
\end{abstract}

\section{Introduction}

Of all nonverbal behaviors---body movement, posture, gaze, proxemics and voice---the face is probably the most commanding and complicated, and perhaps the most confusing. The face, and especially facial movement (or expression), is commanding because it is always visible and therefore always providing information such as emotion and intent \cite{harrigan2008new}. Due to their utility for understanding a human being's mental state, the recognition of automatic facial expressions is becoming a popular field in computer vision.

	Automatic facial expression recognition relies heavily on training datasets. When training a system to recognize facial expressions, the investigator must assume that the training and test data have been accurately labeled. This assumption may or may not be accurate. Traditionally, researchers have categorized facial expressions as expressing happiness, sadness, surprise, fear, anger and disgust. Each of these prototypical facial expressions can be described via a pattern of action units (AUs), based on the Facial Action Coding System (FACS) developed by Ekman and colleagues \cite{ekman2002facial} and supported by Basic Emotion Theory (BET) \cite{russell2011introduction}. 
	
	Early in the process, samples of facial expressions are mostly viewed from the front, collected from actors required to pose preset AUs that readily match prototypical facial expressions \cite{kanade2000comprehensive,lucey2010extended}. Thus, recognition of the emotion can be judged against the ``correct'' expression adopted by the poser. Recently, datasets of spontaneous facial expressions have multiplied \cite{fabian2016emotionet,Mollahosseini2017AffectNet,li2018reliable,li2019blended,zhang2018facial}. This is remarkable progress because the goal of automatic facial expression recognition is to apply to in the real world rather than lab situations (i.e., the ideal world). Facial expressions in real life are ``in the wild'', meaning that they are not necessarily frontal and direct-gazed, nor demonstrate only a very limited number of AU patterns. There may be various combinations of AUs with different gestures, head poses, gazes, and environments.

	Due to the characteristics of in-the-wild facial expressions, applying an emotion label to a certain face can be difficult. AU combinations do not usually fall precisely into the six (or more) prototypical facial expressions \cite{yan2013fast}. EMFACS \cite{kanade2000comprehensive,friesen1983emfacs}, usually understood as a guide for labeling emotions expressed through AU combinations, has become inappropriate in this content. Though previous research has summarized the relationship between AUs and emotion \cite{ekman1969nonverbal,du2014compound}, researchers have only provided general or key AUs for certain emotions. Moreover, the coherence problem has long being a topic of debate in the field of facial expression \cite{DBLP:conf/iccv/RuizWB15,duran2017coherence,DBLP:journals/corr/abs-1910-11111}. Even for the six universally used and prototypical facial expressions, psychologists have not reached a consensus on the relationship between AU patterns and emotions, let alone facial expressions in the wild. In other words, we cannot simply categorize in-the-wild facial expressions by AU pattern, at least not for the moment.
	
		Generally, psychologists have proposed two approaches to studying nonverbal behavior (including facial expressions), either judgement-based or sign-based \cite{harrigan2008new,ekman1969nonverbal}. In judgement-based approaches, observers make inferences about the meaning of facial actions and assign corresponding labels. When classifying facial expressions into a predefined number of emotions or mental activity categories, the agreement of a group of annotators is taken as ground truth, usually by computing the average of the responses of either experts or non-experts. The rationale that judgement-based approaches can provide ``correct'' labels can be explained from an evolutionary perspective: the facial expressions broadcast from a sender (i.e., encoding the signal) should be universally understood by a perceiver (i.e., decoding the signal), or they would be useless and therefore removed. Thus, crowdsourced annotation, a practice that employs many perceivers to heuristically label a target, might be a useful way of labeling in-the-wild facial expressions. As for sign-based approaches, facial motion and deformation are coded into visual classes. Facial actions are then abstracted and described by their location and intensity, such as in FACS. Ideally, a complete description framework would contain all possible perceptible changes that might occur on a face. Most automatic facial expression analysis approaches have attempted to directly categorize facial expressions into basic emotion classes \cite{fasel2003automatic} by AUs, a sign-based approach. 
		
			\begin{figure*}[th]
	\centering
	\includegraphics[width=12cm]{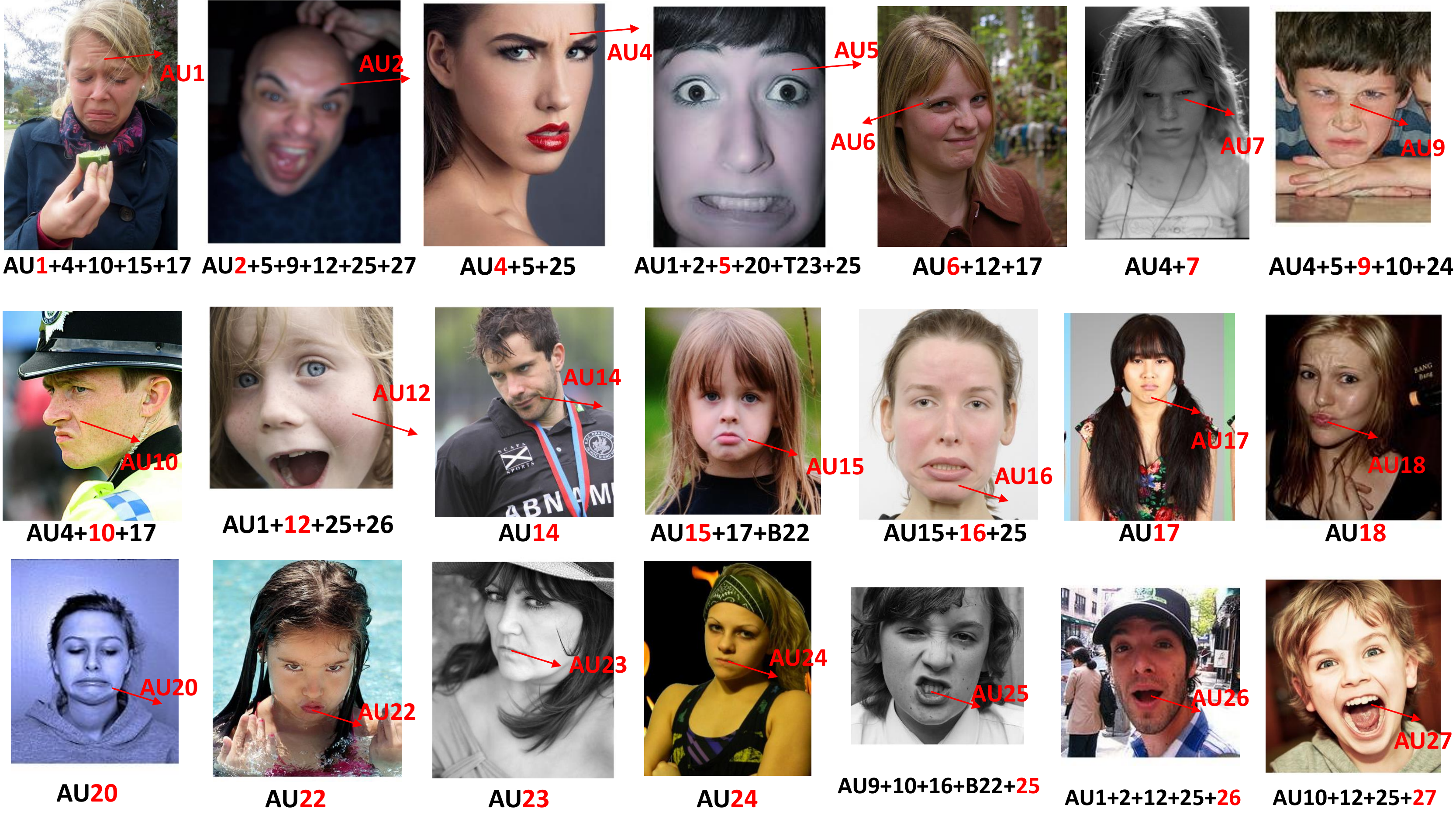}
	\caption{Example samples annotated with typical AUs in the RAF-AU dataset. }
	\label{fg1add}
\end{figure*}

		The judgement-based approach, with subjective estimation, provides the perceivers' judgement because facial expressions are born to be perceived and understood by conspecifics. From this perspective, conspecifics can provide the correct interpretation of a given facial expression. The sign-based approach, with objective description, can tell the computer where and how the facial movements will occur. Ideally,  a qualified mechanism for annotating facial expressions should include both subjective and objective elements. 
		
			With such an understanding of facial expressions and their annotation, the present work provides an updated version of the existing RAF-ML database \cite{li2019blended}, called RAF-AU database\footnote{http://whdeng.cn/RAF/model3.html}, which consists of in-the-wild facial expressions with both subjective (i.e., judgement-based) and objective (i.e., sign-based) annotations. The RAF-ML database contains facial images in different occlusions, illuminations and resolutions collected from the social network and provides multiple labels via crowdsourcing annotations for each facial image. To extend this database with objective elements, we conducted AU coding according to strict criteria by two FACS experts.  Specially, two experienced coders were requested to independently code each the facial image involving 26 kinds of AUs and the inter-coder correlation was 0.6376.
			To our best knowledge, the RAF-AU database is the first facial expression dataset that includes both subjective emotion judgement and objective AU annotations for multi-label expression analysis in-the-wild. Figure \ref{fg1add} shows example images annotated with typical AUs in the RAF-AU dataset.
			Using both crowdsourced and AU annotations, we make the first attempt to investigate the relationship between AUs and perceived emotions in the wild. We explored which AUs contributed the most to each facial expression.
			We then conducted AU detection experiments using popular features and multi-label learning methods. A baseline was provided for AU recognition in the RAF-AU. 
			
\section{Related Work}
Several surveys have offered overviews of facial expression datasets \cite{ko2018brief,li2018deep}. The present research does not repeat their work, but rather inspects how facial movements were annotated with emotion labels. Many of these databases classified facial expressions into types, based on rules such as those in EMFACS (i.e., mapping AUs to facial expressions). Usually, participants were required to pose ``standard'' AUs to express so-called prototypical facial expressions, and those that qualified were included in the database. 
	
In the Cohn-Kanade Expression Database \cite{kanade2000comprehensive}, emotion labels refer to the expression requested rather than what may actually have been performed. AUs for each facial expression (i.e., the apex frame in a sequence) are converted into emotion-specific expressions (e.g., happiness or anger). BU-3DFE \cite{yin20063d} and MMI \cite{pantic2005web} used a similar technique to assign emotion labels to specific facial movements. Researchers have also used this approach to annotate emotions expressed in spontaneous facial expressions. EmotioNet \cite{fabian2016emotionet} is a large-scale database with one million facial expression images collected from the Internet. Most samples were annotated by an automatic AU detection algorithm, and the remaining 10\% were manually annotated using AUs. EmotioNet contains six basic expressions and one neutral expression. The creators defined unique AU patterns that mapped AUs to specific emotions. Therefore, the emotion labels were inferred from the AUs. For example, if the face showed AU4$+$AU15, the emotion label attributed was ``sadness''. Similar mapping rules were used to label compound facial expressions. EmotioNet defined 17 compound expressions, depending on the AU combination. For example, AU4$+$AU20$+$AU25 was defined as fearfully angry. Similarly in \cite{du2014compound}, subjects were required to practice their expressions before each acquisition. They were required to express more than one emotion based on prototypical facial expressions; these were then evaluated by experts for validity. This annotation approach was also based on prototypical facial expression protocol. 

The dimension approach was used to annotate facial expressions from valence (unpleasant to pleasant) and arousal (low to high). For example, AffectNet \cite{Mollahosseini2017AffectNet} contains more than one million images from the Internet that were obtained by querying different search engines using emotion-related tags. Over 450,000 images include manually annotated labels according to eight basic expressions. Another valence-arousal dataset is Aff-Wild \cite{DBLP:journals/ijcv/KolliasTNPZSKZ19,DBLP:conf/cvpr/ZafeiriouKNPZK17} which contains 298 videos with spontaneous facial behaviors collected in arbitrary recording conditions.
Most recently, the Aff-Wild dataset was extended to the Aff-wild2 \cite{DBLP:conf/bmvc/KolliasZ19,DBLP:journals/corr/abs-2001-11409} with more videos and other attribute annotations such as basic expressions and action units. It is by far the largest database providing facial expressions with all three types
of behavior states, i.e., valence and arousal (VA), basic expression and facial action unit.

The situation approach derives emotion labels from the situation (with an anticipated emotion) where facial expressions elicited are taken as a cue. The BP4D-Spontaneous database \cite{zhang2014bp4d} used various tasks in a lab environment to elicit spontaneous facial expressions. For example, smelling an unpleasant odor should be disgusting and thus the facial expression during the task should be one of disgust. The Aff-Wild dataset employed similar elicitation method to record videos containing naturalistic emotional states in arbitrary recording conditions.
		
The self-report approach is another way of annotating facial expression type. The Belfast Induced Natural Emotion Database \cite{sneddon2011belfast} is unique. Recordings are accompanied by self-reports of emotion and intensity, serving as continuous trace-style ratings of valence and intensity. Self-reporting may be the most reliable approach to labeling emotion in facial expressions. Facial movements in this database were also annotated with valence and intensity (or arousal) according to the dimension of the emotion. In Belfast, they used many more words to describe each emotion than the popular six prototypical facial expressions. However, self-reported information is not easily collected and may not reflect the inner mind \cite{barrett2004feelings} and thus not commonly used.
			
Crowdsourcing is a judgment-based approach that labels facial expressions based on human perception. The Real-world Affective Face Database (RAF-DB) \cite{li2018reliable,li2017reliable} is a real-world database containing 29,672 highly diverse facial images downloaded from the Internet. Images were randomly and equally assigned to each labeler, ensuring that there was no direct correlation among the images labeled by a single annotator. For the manually crowdsourced annotations, seven basic (including neutral) and 11 compound emotion labels were applied to the samples. Each image was ensured to be labeled by approximately 40 independent labelers. Later, RAF-ML \cite{li2019blended} was developed. This database contains facial expressions obtained from RAF-DB that offers multiple expressions, and extends the sample collection, 4,908 in total. Another dataset, the Expression in-the-Wild Database (ExpW) \cite{zhang2018facial}, contains 91,793 faces downloaded using Google image searches. Each face image was manually annotated and categorized by human beings into one of seven basic expression categories, but the study didn't explain whether this coding was based on heuristic judgement or mapping rules.

In summary, posed facial expressions are uncommon in real life, so many databases collect spontaneous samples. These spontaneous and in-the-wild facial expressions are very difficult to annotate because they do not usually display commonly accepted and pre-defined AU combinations. Mapping rules comprise the most common approach to labeling facial expression type. There are also dimension, situation, self-report, and crowdsourcing approaches. With images of facial expressions downloaded from the Internet, the subject's emotional state and mindset are unknown, and researchers can't simply infer them by the AU combinations displayed. Therefore, a crowdsourcing approach that leverages human perception to apply annotations is the most suitable of those available. In the present work, we provide a dataset of both judgement-based (i.e., subjective) and sign-based (i.e., objective) annotations. Only a combination of subjective and objective annotations will properly disclose the emotional implications of facial expressions. This is what is provided by RAF-AU.
\section{RAF-AU}
	\subsection{AU annotations}
Facial images from the Internet vary in quality; this is true not only in terms of image resolution, but also spontaneity, since many are posed and thus not emotion-motivated. There are certain individual differences in facial appearance and habitual movement. Some AUs habitually appear, making emotional perception confusing. For example, a smile with AU9 (i.e., a wrinkled nose) is more likely to appear in women, while people with gag teeth are more likely to display AU10 (a raised upper lip). AU coding of facial expressions requires a baseline (or neutral) image, or it is difficult to judge whether certain expressions were due to the face's original appearance or facial movement. For example, eyebrow AU4 can be a permanent feature or transient movement. If we are forced to imagine a neutral face for each face image, then we have to judge whether the expression is caused by a certain AU, perpetual feature, or the influence of another AU. 

In the present research, those with more than two whole faces (305 images in total) were eliminated from the RAF-ML database. There were some images that contained more than two faces. When we checked the voting, we found that participants were not always be voting for the same face. For example, one image contained a smiling face and a sad face, but about 30\% of the participants voted happiness while 25\% voted sadness. After removing these images, 4,601 were left for further coding.

	Two experienced coders independently FACS-coded these face images and arbitrated any disagreement. They usually disagreed on whether a trace of an AU was actually a baseline or emotion-motivated. They also carefully checked and discussed if AUs emerged due to other AUs. The inter-coder correlation (i.e., reliability, see Eq. (\ref{eq1})) was 0.6376. This relatively low reliability was due to the complexity of and lack of clarity in these in-the-wild and blended facial expressions.

\begin{equation}
	\label{eq1}
	R=\frac{2*\#AU(C_1C_2)}{\#All\__{} AU},
	\end{equation}
	where $\#AU(C_1C_2)$ is the number of AUs upon which Coders 1 and 2 agreed and $\#All\__{}AU$ is the total number of AUs in a facial expression scored by two coders. 
	
	\begin{figure}[t]
		\centering
		\includegraphics[width=10cm]{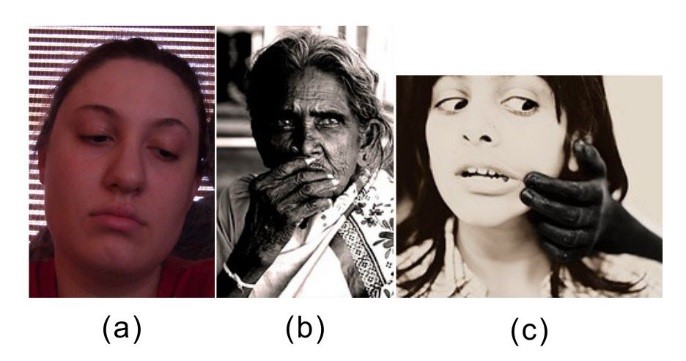}
		\caption{Examples of ``null'' images. }
		\label{fg1}
	\end{figure}

	There were 253 images annotated as null. Some of these facial images didn't show any AUs. Though there were no AUs for these faces, they were still selected for this database because according to perceivers, they expressed emotions. This is not uncommon because perceived emotion is influenced by gaze, head pose, gesture, and facial appearance. Figure \ref{fg1} shows some examples labeled as null. In the figure, (a) is null because the faintly discernible AU17 (i.e., Chin raiser) is not sufficiently obvious; without a baseline, we weren't confident we were avoiding a mistake. In (b), the face of this elderly person has many perpetual traces, and therefore was hard to annotate without a baseline. The AU4 marker seemed locally obvious. However, this is an image of an elderly person that did not show a clear vertical grain between the eyebrows. Therefore, we didn't annotate this image with AU4. The forehead in (c) is covered and the mouth may be stretched by an outer force. These AUs are either invisible or involuntary. 
\subsection{Profile of RAF-AU }
	This database contains 4,601 images obtained from RAF-ML. We have provided AU annotations for all faces. There were 26 AUs used in the annotations, without action descriptors. We list all used AUs and their frequency in Table \ref{tbadd}.
	It should be noted that the AU distribution was largely imbalanced. For example, only a few AU39 (i.e., compressed nostrils) were found in this dataset.
	Some representative examples with frequent AUs in the RAF-AU are shown in Figure \ref{fg1add}.
	\begin{table*}[t]
\renewcommand\arraystretch{1.4}
\setlength{\tabcolsep}{7pt}
\begin{tabular}{|c|c|c||c|c|c|}
		\hline
		\textbf{AU} & \textit{\textbf{Name}}     & \textbf{Num} & \textbf{AU} & \textit{\textbf{Name}} & \textbf{Num}  \\ \hline
		\textbf{1}  & \textit{Inner Brow Raiser} & 1028+(49)    &  \textbf{21} & \textit{Neck Tightener}         & 3     
		    \\ \hline
		\textbf{2}  & \textit{Outer Brow Raiser} & 701+(95)     &   \textbf{22} & \textit{Lip Funneler}           & 196+(148)        \\ \hline
		\textbf{4}  & \textit{Brow Lowerer}      & 1808+(9)     &   \textbf{23} & \textit{Lip Tightener}          & 80+(51)            \\ \hline
		\textbf{5}  & \textit{Upper Lid Raiser}  & 975+(10)     &   \textbf{24} & \textit{Lip Pressor}            & 165+(56)             \\ \hline
		\textbf{6}  & \textit{Cheek Raiser}      & 404+(46)     & \textbf{25} & \textit{Lips part}              & 2830     \\ \hline
		\textbf{7}  & \textit{Lid Tightener}     & 347+(14)     &  \textbf{26} & \textit{Jaw Drop}               & 1089  \\ \hline
		\textbf{9}  & \textit{Nose Wrinkler}     & 749+(25)     &    \textbf{27} & \textit{Mouth Stretch}          & 810     \\ \hline
		\textbf{10} & \textit{Upper Lip Raiser}  & 1274+(116)   & \textbf{28} & \textit{Lip Suck}               & 13+(31) \\ \hline
		\textbf{12} & \textit{Lip Corner Puller} & 1187+(81)    &       \textbf{29} & \textit{Jaw Thrust}          & 27          \\ \hline
		\textbf{14} & \textit{Dimpler}                & 105+(22)     & \textbf{30} & \textit{Jaw Sideways}               & 9  \\ \hline
		\textbf{15} & \textit{Lip Corner Depressor}   & 290+(2)      &  \textbf{32} & \textit{Lip Bite}     & 6  		 \\ \hline
		\textbf{16} & \textit{Lower Lip Depressor}    & 720        & \textbf{33} & \textit{Cheek Blow}            & 1  \\ \hline
		\textbf{17} & \textit{Chin Raiser}            & 541           & \textbf{34} & \textit{Cheek Puff}                 & 9+(1)  \\ \hline	
		\textbf{18} & \textit{Lip Puckerer}          & 118+(11)  &\textbf{35} & \textit{Cheeck Suck} &10+(1) \\\hline
		\textbf{19} & \textit{Tongue Visible}          & 34 & \textbf{39} & \textit{Nostril Compressor}     & 5  	 \\\hline
		    \textbf{20} & \textit{Lip stretcher}          & 199+(33)  &\textbf{43} & \textit{Eyes Closed}            & 148+(9)  \\\hline
		\end{tabular}
\caption{Frequency of all used AUs (without action descriptors) coded by manual FACS coders on the RAF-AU dataset. The numbers in parentheses refer to the sample size of one-sided AUs.}
\label{tbadd}
\end{table*}

	\begin{table*}[tp]
		\renewcommand\arraystretch{1.5}
		\setlength{\tabcolsep}{14pt}
\begin{tabular}{|c|c||c|c||c|c|}
\hline
\multicolumn{2}{|c||}{\textbf{Surprise}}                           & \multicolumn{2}{c||}{\textbf{Fear}}                               & \multicolumn{2}{c|}{\textbf{Disgust}}                            \\ \hline
\textbf{AU25}& 0.5926& \textbf{AU25} & 0.5111&\textbf{AU10}                      & 0.5964  \\ \hline
\textbf{AU5} & 0.4665 & \textbf{AU12} & 0.4033  &\textbf{AU4}                       & 0.5330  \\ \hline
\textbf{AU26} & 0.3820 & \textbf{AU27} & 0.3729 & \textbf{AU17}& 0.2973\\ \hline\hline
\multicolumn{2}{|c||}{\textbf{Happiness}}                          & \multicolumn{2}{c||}{\textbf{Sadness}}                            & \multicolumn{2}{c|}{\textbf{Anger}}                              \\ \hline
\textbf{AU12}                       & 0.7040                      & \textbf{AU4}                       & 0.6723                      & \textbf{AU25}                      & 0.4659                      \\ \hline
\textbf{AU25}                       & 0.5143                      & \textbf{AU25}                      & 0.3462                      & \textbf{AU9}                       & 0.4337                      \\ \hline
\textbf{AU27} & 0.2491& \textbf{AU1} & 0.2979 & \textbf{AU10}& 0.4236\\ \hline
\end{tabular}
		\caption{AUs that contribute most (variance) for these facial expressions in RAF-AU.} 
		\label{tb1}
\end{table*}

	\begin{table}[htp]
		\renewcommand\arraystretch{1.5}
		\setlength{\tabcolsep}{7pt}
		\begin{tabular}{|c|c|c|c|c|c|c|c|}
			\hline
			& \textbf{Sur} & \textbf{Fea} & \textbf{Dis} & \textbf{Hap} & \textbf{Sad} & \textbf{Ang} & \textbf{Neu} \\ \hline
			\textbf{Sur}  & 1.0000 & 0.6802  & 0.1733 & 0.4908 & 0.3284 & 0.3688 & -0.3784 \\ \hline
			\textbf{Fea}      & 0.6802 & 1.0000 & 0.3645 & 0.8168 & 0.5754 & 0.7665 & -0.8262  \\ \hline
			\textbf{Dis}   & 0.1733 & 0.3645  & 1.0000  & 0.2497 & 0.6997  & 0.6266  & -0.4654 \\ \hline
			\textbf{Hap} & 0.4908  & 0.8168 & 0.2497 & 1.0000  & 0.5048 & 0.6207  & -0.7539 \\ \hline
			\textbf{Sad}   & 0.3284 & 0.5754  & 0.6997  & 0.5048   & 1.0000  & 0.6342 & -0.4866 \\ \hline
			\textbf{Ang}     & 0.3688  & 0.7665 & 0.6266 & 0.6207  & 0.6342 & 1.0000   & -0.8170  \\ \hline
			\textbf{Neu}   & -0.3784 & -0.8262 & -0.4654  & -0.7539 & -0.4866  & -0.8170  & 1.0000 \\ \hline
		\end{tabular}
		\caption{The relationship of the facial expressions based on the AU combination for RAF-AU. Sur=Surprise, Fea=Fear, Dis=Disgust, Hap=Happiness, Sad=Sadness, Ang=Anger, Neu=Neutral.}
		\label{tb2}
	\end{table}

	\begin{table}[htp]
	\renewcommand\arraystretch{1.5}
	\setlength{\tabcolsep}{7pt}
	\begin{tabular}{|c|c|c|c|c|c|c|c|}
		\hline
		& \textbf{Sur} & \textbf{Fea} & \textbf{Dis} & \textbf{Hap} & \textbf{Sad} & \textbf{Ang} & \textbf{Neu} \\ \hline
		\textbf{Sur}  & 1.0000 &0.2706 &-0.5585 &	0.1544 &	-0.4168 &	-0.3756 &	-0.2598  \\ \hline
		\textbf{Fea} & 0.2706 &1.0000 &	-0.4242 &	-0.2193 &	-0.1224 &	-0.1628 &	-0.3568   \\ \hline
		\textbf{Dis}&-0.5585 & -0.4242 	&1.0000 &	-0.2673 &	0.0686 &	0.0739 	& 0.1447  \\ \hline
		\textbf{Hap} & 0.1544 &-0.2193 	&-0.2673 &	1.0000 &-0.3023 &-0.2986 &	-0.0387  \\ \hline
		\textbf{Sad} &-0.4168 &-0.1224 &	0.0686 & -0.3023 &	1.0000 &	-0.2060 &	0.0817  \\ \hline
		\textbf{Ang} & -0.3756 	&-0.1628 &	0.0739 &	-0.2986 &	-0.2060 &	1.0000 &	-0.1858   \\ \hline
		\textbf{Neu} & -0.2598 &-0.3568 &	0.1447 &	-0.0387 &	0.0817 &	-0.1858 &	1.0000  \\ \hline
	\end{tabular}
	\caption{The correlation coefficients between the facial expressions for RAF-AU. Sur=Surprise, Fea=Fear, Dis=Disgust, Hap=Happiness, Sad=Sadness, Ang=Anger, Neu=Neutral.}
	\label{tb3}
\end{table}
	Among these AU annotations, some were one-side AUs. Facial action units are not always the same for left and right halves of faces; they may also vary in intensity. One-sided AUs were labeled as L or R, depending on which half of the face acted. Among the 4,601 faces, we found 219 right-only and 304 left-only AUs. For the unweighted condition (the basic unit was facial expression, and it did not matter if it contained one or more one-sided AUs), we found 163 right-only and 232 left-only AUs. This finding suggests that the left side of the face is more expressive than the right. Psychologists may be interested in these types of characteristics and thus should further investigate this topic. In addition to the left- and right-only AUs, some AUs only contained top or bottom elements. For example, AU23 indicates tightened lips, and sometimes this is only expressed by half (top or bottom). For these instances, we annotated T for top only and B for bottom only. There were 131 T and 179 B faces in RAF-AU.
	
	By annotating for both AU and perceived emotion, we were able to explore which AUs contributed the most to blended emotions. Through a linear transformation, the AUs and perceived emotions were made equivalent in the same space. That is, the AUs of each image were linearly transformed to obtain the expression corresponding to the image. 
	Through matrix transformation, we will be able to get the linear relationship between AUs and perceived emotions. Accordingly, the AUs of each image can be linearly transformed to obtain the expression possibilities corresponding to the image.
	Table \ref{tb1} shows the three AUs contributing the most to each facial expression. We explored the inner relationship of the expression itself by using the expression matrix and calculating the relationship of the facial expression based on the AU combination (see Table \ref{tb2}). In addition, we provide the correlation coefficients for the facial expressions in RAF-AU (see Table \ref{tb3}).

	\section{Baseline Evaluation on RAF-AU}
	\subsection{Pre-processing and dataset split}
	We first filtered out all 253 images annotated with ``null'' as they may contain irrelevant information which would largely distract the AU detection.
	Then, we chose 13 kinds of AUs that occur more than 8\% base rate in RAF-AU dataset for experiment and analysis. Specifically, each image was annotated +1 or -1 if an AU is present or absent, and 0 for one-sided AUs.
	During pre-processing, manually annotated five facial landmarks provided in RAF-ML \cite{li2019blended} were employed to register all images to a reference face using an affine transformation, resulting 100*100 cropped images. Then gray-scale samples were transformed for the following feature extraction.
	For dataset split, we divided RAF-AU into training part and test part, where the size of the training part is four times larger than the test part.
	
	\subsection{Feature extraction and classification}
	For the comparison purpose, we implemented two handcrafted features and three deep learning features.
	For hand-crafted feature extraction, we have tried histogram of orientated gradients (HOG)~\cite{dalal2005histograms} and Local binary patterns (LBP)~\cite{ojala2002multiresolution}. For HOG, we used this shape-based segmentation dividing the image into 10*10 pixel blocks of four 5*5 pixel cells with no overlapping. By setting 10 bins for each histogram, we got a 4000-dimensional feature vector per aligned image. For LBP, we selected the 59-bin $LBP_{8,2}^{u2}$ operator, and divided the 100*100 pixel images into 100 regions of 10*10 grid size, which was empirically found to achieve relatively good performance for expression classification.

	For deep learning feature extraction, we first employed the already trained baseDCNN and DBM-CNN provided in \cite{li2019blended}, then 2000-dimensional deep features learned from raw data can be extracted from the penultimate fully connected layer of these two DCNNs for both training set and test set in RAF-AU. 
	We also tried a multi-label CNN for AU detection. Let us assume that there are $k$ AU categories and $n$ training samples. Given a set of ground truth label $\boldsymbol{y} \in \{-1,+1,0\}^k$ and the corresponding prediction results $\boldsymbol{p} \in \mathbb{R}^k$ for all $k$ AU labels, the goal is to minimize the following multi-label cross entropy loss:
	\begin{equation}
	L=\frac{-1}{n}\sum_{i=1}^{n}\sum_{j=1}^{k} 
	\left\{
	[y_i^j \textgreater 0]\log p_i^j+ [y_i^j \textless 0]\log (1-p_i^j) 
	\right\},
	\end{equation}
	where $[ \cdot ]$ is an indicator function that returns 1 if the Boolean expression is true, and 0 otherwise. And $y_i^j$ is the ground truth for the $i$-th sample of $j$-th AU, $p_i^j$ is the predicted probability for the $i$-th sample of $j$-th AU. We then also extracted the output of the penultimate fully connected layer as the final feature representation, resulting in 2000-dimensional vectors. Table \ref{deep_base} displays  detailed network architectures  of  this AU-CNN trained on the RAF-AU dataset.

	\begin{table*}[tbp]
\renewcommand{\arraystretch}{1}
   \setlength{\tabcolsep}{7pt}
\centering
\begin{tabular}{@{}|lccccccccc@{}|}
\hline
\multirow{2}{*}{\begin{tabular}[c]{@{}c@{}}Layer \\ Type\end{tabular}} & 1    & 2    & 3     & 4    & 5    & 6     & 7    & 8    & 9      \\ 
& Conv & ReLu & MPool & Conv & ReLu & MPool & Conv & ReLu & Conv \\ \hline
Kernel  & 3    & -    & 2     & 3     & -    & 2     & 3        & -    & 3              \\
output  & 64   & -    & -     & 96   & -    & -     & 128  & -    & 128   \\
Stride  & 1    & 1    & 2     & 1    & 1    & 2     & 1      & 1    & 1                \\
Pad & 1    & 0    & 0     & 1    & 0    & 0     & 1      & 0    & 1        \\ \hline\hline
\multirow{2}{*}{\begin{tabular}[c]{@{}c@{}}Layer \\ Type\end{tabular}}    & 10   & 11    & 12   & 13   & 14   & 15   & 16      & 17   & 18    \\ 
& ReLu & MPool & Conv & ReLu & Conv & ReLu &  FC   & ReLu &  FC \\ \hline
Kernel     & -    & 2     & 3     & -    & 3        & -    &            & -    &            \\
output    & -    & -     & 256  & -    & 256  & -    & 2000 & -    &          13  \\
Stride      & 1    & 2     & 1    & 1    & 1      & 1    &            & 1    &             \\
Pad  & 0    & 0     & 1    & 0    & 1      & 0    &           & 0    &  \\ \hline
\end{tabular}
\caption{The network configuration parameters in the AU-CNN. }
\label{deep_base}
\end{table*}

	During AU detection, support vector machine with linear kernel  implemented in LIBSVM \cite{CC01a} was utilized for the one-versus-all binary classification.  Given a training set $\{(x_i,y_i), i=1,\dots,n\}$, where $x_i \in \mathbb{R}^d$ and $y_i \in \{+1,-1\}$ (samples with one-sided AU has been omitted during this AU detection), then the testing sample can be detected by optimizing: 
	\begin{equation}
	\min_{w}\frac{1}{2}||w||^2+C\sum_{i=1}^{n}max(1-y_iw^Tx_i,0).
	\end{equation}
	The penalty parameter $C$ of SVM was fixed to 1 for all different features.
	
			\begin{table*}[t]
		\renewcommand\arraystretch{1.35}
		\setlength{\tabcolsep}{7.5pt}
		\begin{tabular}{|p{0.08\textwidth}<{\centering}|p{0.08\textwidth}<{\centering}|p{0.09\textwidth}<{\centering}|p{0.09\textwidth}<{\centering}|p{0.11\textwidth}<{\centering}|p{0.125\textwidth}<{\centering}|p{0.11\textwidth}<{\centering}|}
\hline
AUs         & Number & HOG & LBP & BaseDCNN & DBM-CNN & AU-CNN \\ \hline
\textbf{1}  &   1028     &  76.83   &  78.72   &   72.31       &    71.74     &82.03        \\ \hline
\textbf{2}  &     701   &    84.67 &  87.76   &    83.54      &  80.67       &  90.49      \\ \hline
\textbf{4}  &     1808   &  72.95   &  74.27   &   75.67       &  78.73       &        86.28\\ \hline
\textbf{5}  &      975  &  78.19   &   76.43  &   83.25       &   87.84      & 88.86       \\ \hline
\textbf{6}  &      404  &    84.61 &   80.68  &  81.98        &   84.29      &87.41        \\ \hline
\textbf{9}  &  749      &   84.09  &   88.49  &    86.79      &   86.76      &  90.03      \\ \hline
\textbf{10} &    1274    &  78.98   &  76.63   &      80.85    &   81.56      &87.89        \\ \hline
\textbf{12} &    1187    &  80.51   &   82.27  &   86.23      &   86.17      &88.48        \\ \hline
\textbf{16} &   720     &  78.47   &  77.84   &  77.71        &   79.67      &86.11        \\ \hline
\textbf{17} &    541    &   81.81  &  79.55   &   80.07       &  78.29       &        88.58\\ \hline
\textbf{25} &     2830   &  85.13   &  86.24   &  89.87        &    91.18     &95.33        \\ \hline
\textbf{26} &    1089    &   66.27  &  67.72   &   75.12       &     78.82    &86.31        \\ \hline
\textbf{27} &   810     &   91.48  &  93.11   &     93.42     &   93.65      & 95.77       \\ \hline
AVG           & -            &  80.31   &  80.75   &   82.06       &  83.03       &      \textbf{88.73}   \\ \hline
\end{tabular}
		\caption{Performances of the AUC-ROC for AU detection on RAF-AU dataset using different features.}
		\label{tb4}
\end{table*}

	\subsection{Evaluation metrics}
	In terms of evaluating the performance on AU detection, two different metrics were employed: the area underneath the receiver-operator characteristic (ROC) curve (AUC-ROC) and the F1 score.
	The rank metric ROC curve visualizes the trade-off between sensitivity and specificity by plotting both values as a function of a varying classification threshold. And the threshold metric F1 score is defined as $F1=\frac{2RP}{R+P}$, where $R$ and $P$ denote recall (the number of correctly recognized samples divided by the actual number of all samples with the target AU) and precision (the number of correctly recognized samples divided by the total number of samples detected with the target AU), respectively. We then computed the average over all 13 AUs (AVG) to measure the overall performance.
	
	\subsection{AU detection results}
	In Table \ref{tb4} and Table \ref{tb5}, we show the AUC-ROC and F1 score results for each of the 13 AUs in RAF-AU using four different features, respectively. We also list the statistic regarding the AU occurrence, i.e., the number of positive samples for each AU. It can be seen that the AU distribution in RAF-AU is more imbalanced than those lab-controlled datasets.
	With regard to individual AU detection, our baselines can yield relatively good performance on frequently occurring AUs (e.g., AUs 25, 27, 5, 4 and 12). However, we also observe a significant drop in the performance on other less common AUs.
	When comparing different features, the deep learning feature AU-CNN can achieve comparable and better AU detection rates in terms of both average AUC and F1 score.
	Nevertheless, when compared to the accuracy achieved on other lab-controlled datasets, there is still room for improvement on this challenging realistic dataset which contains various naturalistic illuminations, occlusions, hear poses and obvious imbalanced distribution.

			\begin{table*}[t]
		\renewcommand\arraystretch{1.35}
		\setlength{\tabcolsep}{7.5pt}
		\begin{tabular}{|p{0.08\textwidth}<{\centering}|p{0.08\textwidth}<{\centering}|p{0.09\textwidth}<{\centering}|p{0.09\textwidth}<{\centering}|p{0.11\textwidth}<{\centering}|p{0.125\textwidth}<{\centering}|p{0.11\textwidth}<{\centering}|}
			\hline
			AUs         & Number & HOG & LBP & BaseDCNN & DBM-CNN & AU-CNN \\ \hline
			\textbf{1}  &    1028    &   33.70  &    50.00 &   29.60       & 35.42        &  60.47      \\ \hline
			\textbf{2}  &      701  &   45.00  &   50.23  &   38.50       &32.49         &    65.59    \\ \hline
			\textbf{4}  &    1808    &  58.86   &  57.89   &    62.35      &  65.47       &  73.44      \\ \hline
			\textbf{5}  &       975 &   45.43  &    37.29 &     51.53     &59.52         & 69.69       \\ \hline
			\textbf{6}  &   404   &  26.09     &  26.80   &  17.78        &18.18         &  58.21      \\ \hline
			\textbf{9}  &     749   &   52.94  &  59.29   &      52.24    & 51.48        &    67.44    \\ \hline
			\textbf{10} &  1274      & 49.30    & 49.65    &    49.52      &         51.99&    68.41    \\ \hline
			\textbf{12} &   1187     &  54.72   & 60.29    &    64.76      & 62.09        &  69.62      \\ \hline
			\textbf{16} &   720     &    21.65 &   27.03  &      25.37    &    31.63     &   59.38     \\ \hline
			\textbf{17} &    541    &7.27     & 20.74    &     18.60     &  16.26       &   25.64    \\ \hline
			\textbf{25} &     2830   &   85.07  &  85.16   &  87.50        & 88.87        &   92.14     \\ \hline
			\textbf{26} &    1089    & 20.00    &   17.90  &      42.11    &48.84         &   64.65     \\ \hline
			\textbf{27} &   810     &   68.69  &  72.11   &  70.99        & 71.23        &  82.67      \\ \hline
			AVG         & -      & 43.75    &    47.26 &    46.99      &      48.73     & \textbf{65.95}        \\ \hline
		\end{tabular}
			\caption{Performances of the F1 score for AU detection on RAF-AU dataset using different features.}
	\label{tb5}
\end{table*}

	\section{Conclusions}
	In real life, facial expressions occur in the wild, and thus their emotional meaning cannot be absolutely defined. The crowdsourcing approach provides a subjective label based on human perception. Thus, we update the previous RAF-ML database by providing AU coding and removing certain confusing samples. The present database, RAF-AU, provides both AU- and judgement-based annotations from objective and subjective approaches, respectively. Thus, this database provides a fuller picture of given facial expressions. Based on these annotations, we were able to investigate the relationship between objective description and subjective understanding. We also provide a set of baselines for RAF-AU depending on different features. And the deep learning feature obtained using a multi-label AU detection CNN achieve the best detection rate. Further research should be conducted to study this relationship, with more samples and across various cultures.  

\section{Acknowledgement}
This work was partially supported by the National Natural Science Foundation of China under Grants No. 61871052. We thank Yong Ke, Huaxiu Li for their assistance in AU coding.
\bibliographystyle{splncs}
\bibliography{egbib}

\end{document}